\documentclass[10pt,twocolumn,letterpaper]{article}

\usepackage{cvpr}              % To produce the CAMERA-READY version

\usepackage[table,xcdraw]{xcolor}
\usepackage{bm}
\usepackage{booktabs}
\usepackage{arydshln}
\usepackage{times}
\usepackage{epsfig}
\usepackage{graphicx}
\usepackage{amsmath}
\usepackage{amssymb}
\usepackage{caption}

\usepackage{array}
\usepackage[table,xcdraw]{xcolor}
\usepackage{bm}

\newcommand\blfootnote[1]{%
  \begingroup
  \renewcommand\thefootnote{}%
  \footnote{#1}%
  \addtocounter{footnote}{-1}%
  \endgroup
}

\definecolor{cvprblue}{rgb}{0.21,0.49,0.74}
\usepackage[pagebackref,breaklinks,colorlinks,citecolor=cvprblue]{hyperref}

\title{Deep CNN Face Matchers Inherently Support Revocable Biometric Templates}

\author{Aman Bhatta \quad\quad  Michael C. King \quad\quad Kevin W. Bowyer${\thanks{Dr. Bowyer is a member of the FaceTec (\url{facetec.com}) Advisory Board.  Results in this paper do not necessarily relate to FaceTec products.}}$ \\\\
University of Notre Dame\\
Florida Insitute of Technology\\
{{\tt\small \{abhatta,kwb\}@nd.edu,michaelking@fit.edu}}}

\begin{document}
\maketitle  

%%%%%%%%% ABSTRACT
\begin{abstract}
One common critique of biometric authentication is that if an individual’s biometric is compromised, then the individual has no recourse.  The concept of revocable biometrics was developed to address this concern.  A biometric scheme is revocable if an individual can have their current enrollment in the scheme revoked, so that the compromised biometric template becomes worthless, and the individual can re-enroll with a new template that has similar recognition power.  We show that modern deep CNN face matchers inherently allow for a robust revocable biometric scheme.  For a given state-of-the-art deep CNN backbone and training set, it is possible to generate an unlimited number of distinct face matcher models that have both (1) equivalent recognition power, and (2) strongly incompatible biometric templates.  The equivalent recognition power extends to the point of generating impostor and genuine distributions that have the same shape and placement on the similarity dimension, meaning that the models can share a similarity threshold for a 1-in-10,000 false match rate.  The biometric templates from different model instances are so strongly incompatible that the cross-instance similarity score for images of the same person is typically lower than the same-instance similarity score for images of different persons.  That is, a stolen biometric template that is revoked is of less value in attempting to match the re-enrolled identity than the average impostor template.
We also explore the feasibility of using a Vision Transformer (ViT) backbone-based face matcher in the revocable biometric system proposed in this work and demonstrate that it is less suitable compared to typical ResNet-based deep CNN backbones. \blfootnote{\textit{Code at: https://github.com/abhatta1234/Revocable-Biometrics}}

\end{abstract}

%%%%%%%%% BODY TEXT
\section{Introduction}

Biometric templates are traditionally considered irreplaceable; once stolen, they are thought to be permanently compromised. This misunderstanding is even ingrained in legislation. For example, the Biometric Information Privacy Act (740 ILCS 14/5 Section 5c) \cite{BIPA}, enacted by the state of Illinois in the United States, states, ``Biometrics are unlike other unique identifiers that are used to access finances or other sensitive information. For example, social security numbers, when compromised, can be changed. Biometrics, however, are biologically unique to the individual; therefore, once compromised, the individual has no recourse, is at heightened risk for identity theft, and is likely to withdraw from biometric-facilitated transactions" \cite{BIPA}.  An article discussing fingerprints states, ``Biometric data might provide a way to identify people with a high degree of accuracy, but once it is stolen, there is nothing you can do to make it secure again. Of course, if your fingerprint is stolen, you could always use another finger, but you could only do this 10 times. … If enough people have their biometric data exposed, eventually some systems could become unusable because so many users won’t be able to securely log in to them " \cite{theconversationStolenFingerprints}. Similar ideas are discussed in these articles \cite{spiceworksRealRisks,nbcnewsBiometricScanning}.

\begin{figure}[t]
    \centering
    \includegraphics[width=\columnwidth]{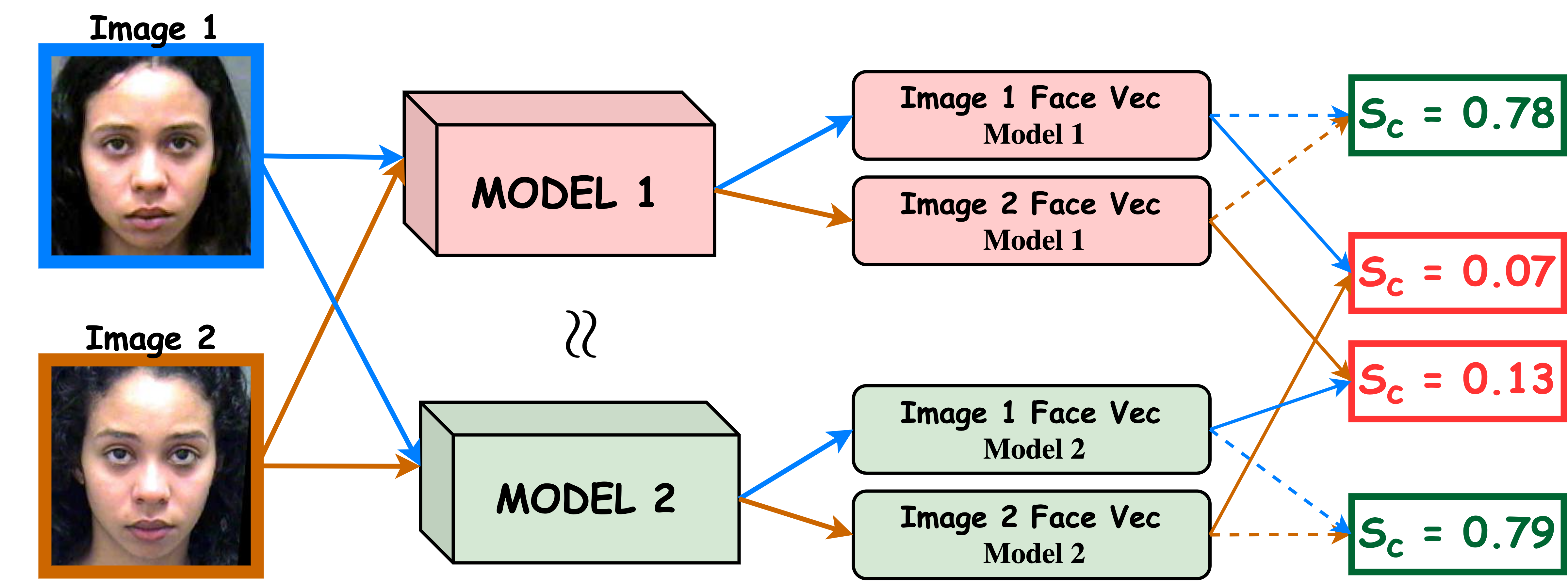}
    \caption{Cosine Similarity scores for genuine pairs using feature vectors from Model 1 and Model 2, {\bf both identical end-to-end models but trained separately}, demonstrates impostor-like behavior.
    % This holds true for N different instances of the same model. Note that any cross-model comparison of genuine pairs, or even the same image across different instances, results in low similarity scores. 
    [Key - $S_c$: Cosine similarity]}
    \label{fig:intro_fig}
    \vspace{-0.75em}
\end{figure}

The notion of ``biometric identity" being stolen, as suggested in various writings, is somewhat misleading. When a biometric sample, such as a face, fingerprint, or iris, is used, the ``identity" refers to the biometric template (also known as the feature vector or embedding) generated by the specific model in use. In cases where the ``biometric identity" is compromised, two potential breaches can occur: either at the image level used for enrollment or at the feature extraction level, where the embedding vector is compromised. In the context of face recognition applications, there is a general expectation that individuals' face images are already publicly available, making a breach at the image level less pressing. As shown in Figure \ref{fig:intro_fig}, the match score between two genuine samples from different instances of the same end-to-end model results in an impostor score. Consequently, replacing the model used for a given identity renders the previously enrolled template unusable for future verifications, allowing breaches involving the enrolled feature template to be easily mitigated.

This work addresses verification tasks within the framework of revocable biometrics, where compromised biometric templates can be revoked and replaced with new ones to ensure security. In the proposed system, where multiple instances of matchers are required, it is essential that each matcher performs equivalently to ensure consistent recognition accuracy after re-enrollment. 
We demonstrate that $N$ distinct matchers with similar performance can be produced by training multiple instances of the same end-to-end model. 
It is also critical that a past template becomes ineffective once revoked.
We demonstrate this showing that cross-model genuine pairs produce scores no better than a typical impostor pair. 
This ensures the revoked template cannot be used to impersonate the legitimate user after re-enrollment.

Major contributions of our work include: 
\begin{itemize}
    \item We propose a new framework for revocable biometrics that utilizes different instances of the same trained end-to-end models, utilizing the inherent non-linear transformations in ResNet-based deep CNNs to generate revocable templates.
    \item We explore the feasibility of a Vision Transformer-based backbone (ViT) within the proposed revocable biometrics framework and demonstrate that Vision Transformer networks are less suitable for this framework.
\end{itemize}

\section{Literature Review}
In this section, we review popular techniques for cancellable or revocable biometrics. For more detailed descriptions, we refer readers to \cite{manisha_air_2020,patel_ieee_2015}. One of the earliest and most well-known approaches to cancellable biometrics is the use of non-invertible transformations, a widely recognized method for generating cancellable biometric templates. The core idea is to apply linear or non-linear non-invertible transformations, such as Cartesian, polar, or functional transforms, to the biometric data during enrollment \cite{bolle_pr_2002,chikkerur_btas_2008,ratha_tpami_2007,ratha_ibm_2001}. While these methods are simple and effective, they have several limitations. For example, they are vulnerable to small changes in the signals and can become unstable near sharp boundary points. Another commonly used non-invertible method is random projection \cite{jin_prl_2014,kaur_mta_2017,kim_btas_2007,lee_icpr_2018,pillai_acssp_2010,pillai_tpami_2011}. In this approach, the extracted features are projected onto a random subspace, typically smaller than the original feature space, and matching is performed in this reduced subspace \cite{achlioptas_jcss_2003}.

Another well-known approach, largely inspired by the success of convolutional networks, is the use of cancellable biometric filters \cite{savvides_icpr_2004}. The core idea is to encrypt biometric templates using a user-specific random convolution kernel during training. This random convolution kernel functions as a personal identification number (PIN). The convolved training images are then used to generate a minimum average correlation energy (MACE) biometric filter. This encrypted filter is stored and used for authentication. An extension of the random projection approach is BioHashing, which builds on similar principles. In BioHashing \cite{connie_ipl_2005,jin_pr_2004,kong_pr_2006,leng_neuro_2013,teoh_pr_2008,teoh_smcc_2007,teoh_tpami_2006}, a feature extraction method, such as wavelet transform, is first applied to extract biometric features from the input biometric data. Using a user-specific tokenized random number (TRN), a set of orthogonal pseudo-random vectors is generated. The dot product between the feature vector and these random vectors is then computed. Finally, binary discretization is applied to generate the BioHash template. The BioHashing framework is designed as a one-way transform, offering a high level of security for both the biometric data and external factors. Another convolution-based approach, known as bio-convolving, is used for biometrics where templates can be represented as a set of sequences, such as in online signature verification \cite{maiorana_smcp_2010}. A popular alternative is the Bloom filter-based approach, which utilizes a space-efficient probabilistic data structure to support membership queries \cite{rathgeb_cs_2014,rathgeb_icb_2013,rathgeb_ietbiom_2014}. Other techniques, including random perturbations \cite{zuo_icpr_2008}, salting methods \cite{zuo_icpr_2008}, and hybrid approaches \cite{boult_fgr_2006,boult_cvpr_2007}, are also employed to secure biometric templates. Several other cancellable biometrics schemes have been developed in \cite{mtibaa_atsip_2018,raja_icmew_2018,raja_fusion_2018,saito_joint_2016}. A neural network-based cancellable biometric scheme was first proposed in \cite{talreja_globalsip_2017}. More recently, researchers have been exploring the use of homomorphic encryption for privacy preservation in biometric applications \cite{yalavarthi_fg_2024}. For further details, refer to the survey in \cite{yang_sensors_2023}.

Our revocable framework is inspired by cancellable biometric filters; however, instead of creating a user-specific filter, we utilize \(N\) sets of equally capable models to revoke the enrollment of identity(ies) in the event of a stolen template or if the user wishes to perform a new enrollment. Our proposed framework also adheres to the ISO/IEC International Standard 24745 \cite{ISO24745}, which provides guidance on protecting individual privacy and outlines four key properties for cancellable biometric templates: non-invertibility/irreversibility, revocability, unlinkability/non-linkability, and performance preservation.

\noindent{\bf (a) Irreversibility} – The feature template must be computationally infeasible to reverse in order to reconstruct the original biometric data. Recently, a research field has emerged focused on recreating biometric identities from feature vectors, but this requires precise knowledge of the model's training process, architecture details, and other specific information. Furthermore, deep neural network models consist of stacked, irreversible non-linear transformation functions, making it difficult to regenerate biometric identities from feature templates produced by such models.

\noindent{\bf (b) Unlinkability} – The protected samples should be unlinkable, meaning it should be challenging to correlate a person's biometric data across different databases. Similar to irreversibility, without detailed knowledge of the training process, linking the generated template to the original model is highly difficult. 

\noindent{\bf (c) Revocability} – The template can be revoked in the event of a breach. Given \(N\) models, it is straightforward to revoke a previous template and generate a new one using a different model within our framework. 

\noindent{\bf (d) Performance preservation} – Our use of \(N\) different instances of the same end-to-end model within our framework requires that each instance maintains similar performance, and we demonstrate that generating such \(N\) models is feasible.

Finally, from an application standpoint, we should also consider the following:

\noindent{\bf (e) Ease of enrollment} – In the event of a template breach, the compromised template can be revoked and replaced by generating a new model for the specific identity without compromising accuracy. This process does not require re-enrollment of the entire gallery of templates.

\section{N Distinct Models with Equivalent Accuracy}
Consider a scenario with \(N\) instances of equally capable models, each obtained through identical end-to-end training using the same backbone, loss function, dataset, and hyperparameter configurations. Our core hypothesis for the \(N\)-model Revocable Biometrics framework is two-fold: a) We can generate \(N\) instances of the model that are equally accurate, i.e., there is no significant performance variation across the \(N\) trained instances, and b) any cross-model genuine-pair comparisons will result in impostor-level scores. However, for this system to function effectively, two key considerations must be verified:
\begin{itemize} 
\item Is it possible to train \(N\) distinct models that achieve near-identical accuracy? 
\item Does cross-matching for genuine pairs across these \(N\) distinct models result in very low similarity scores in cross-model comparisons? 
\end{itemize}

\subsection{Models With Same Impostor and Genuine Distributions} \label{nmodels}
To demonstrate that training \(N\) distinct yet capable models is a feasible task, we begin by training 10 models with three different backbones: ResNet-18, ResNet-100, and ViT, using ArcFace loss and identical hyperparameter configurations. We present two metrics to show that these models are equally capable and similarly trained: a) Average 1:1 Verification Accuracy (\%), including standard deviations across trained models, and b) Average d-prime, along with standard deviations across trained models.

\noindent{\bf (a) 1:1 Verification Accuracy.} We follow the standard 1:1 Verification Accuracy (\%) protocol and report the average accuracy across five standard face recognition benchmarks: LFW \cite{lfw}, CFP-FP \cite{cfpfp}, AGEDB-30 \cite{agedb30}, CALFW \cite{calfw}, and CPLFW \cite{cplfw}, providing a more generalized measure of model's overall performance. The image pairs in this dataset represent ``in the wild" test scenarios. %Benchmark accuracy serves as a global measure of the model's overall performance.

\noindent{\bf (b) d-prime metric.} We adopt the approach outlined in \cite{d_prime_formula}. While 1:1 Verification Accuracy (\%) offers a general performance evaluation, it is crucial to ensure that any biometric system performs equitably across different racial groups. To verify that the \(N\) distinct trained models exhibit consistent performance across racial groups, we present d-prime evaluations for four demographic groups: Caucasian males (C M), Caucasian females (C F), African-American males (AA M), and African-American females (AA F), as represented in the MORPH dataset. The image pairs in this dataset represent ``controlled" or ``ID quality" test scenarios.

\setlength\extrarowheight{10pt}
\begin{table}[!h]
\centering
\resizebox{1\columnwidth}{!}{%
\begin{tabular}{c|c|cccc}
           & Accuracy (\%) &     & d-prime &     &      \\  \hline
 Backbone  & Benchmark Avg.                 & C F & AA F    & C M & AA M \\ \hline
 R18       &  96.41 $\pm$ {\bf 0.07}                              & 6.27 $\pm$ {\bf 0.02}   & 6.43 $\pm$ {\bf 0.02}   &  7.00 $\pm$ {\bf 0.03}  &  7.53 $\pm$ {\bf 0.01}   \\
 R100      & 97.49 $\pm$ {\bf 0.04}                              & 8.08 $\pm$ {\bf 0.03}   &  8.41 $\pm$ {\bf 0.03}      & 8.93 $\pm$ {\bf 0.04}   & 9.84 $\pm$ {\bf 0.02}    \\ 
ViT &     96.95 $\pm$ {\bf \textcolor{red}{0.19}}                          &  6.52 $\pm$ {\bf \textcolor{red}{0.24}}  &  6.54 $\pm$ {\bf \textcolor{red}{0.25}}       & 7.38 $\pm$ {\bf \textcolor{red}{0.26}}   & 7.56 $\pm$ {\bf \textcolor{red} {0.26}}     \\ \hline
\end{tabular}
}
\vspace{0.1em}
\caption{ {\bf Benchmark accuracy and d-prime w/ standard deviation}. For ResNet-based networks, the standard deviations in both benchmark accuracy and d-prime across all demographic groups are small, indicating consistent performance across all 10 trained instances. However, for ViT, there is significant variation in performance across the 10 instances, making it a less reliable choice for the backbone compared to the more stable ResNet networks. }
\label{tab:n-models}
\end{table}

\begin{figure*}[t]
\centering
  \begin{subfigure}[b]{1\linewidth}
    \centering
      \begin{subfigure}[b]{0.33\linewidth}
        \centering
          \includegraphics[width=1\linewidth]{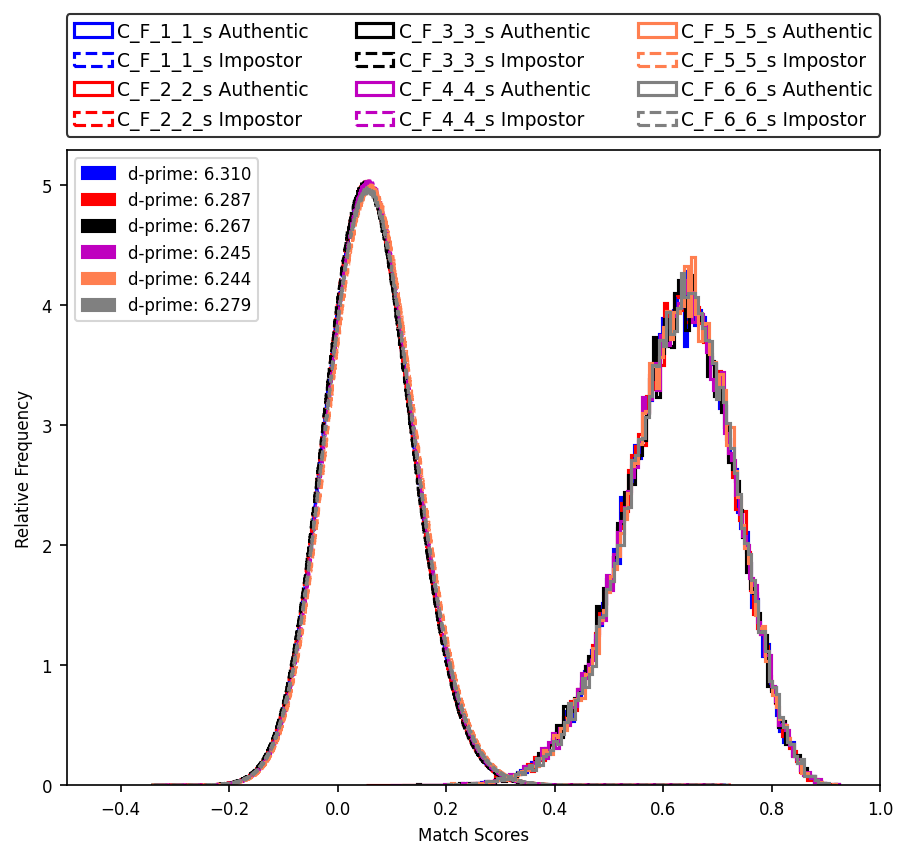}
          \caption{Resnet18}\label{r18}
      \end{subfigure}
      \hfill
      \begin{subfigure}[b]{0.33\linewidth}
        \centering
          \includegraphics[width=1\linewidth]{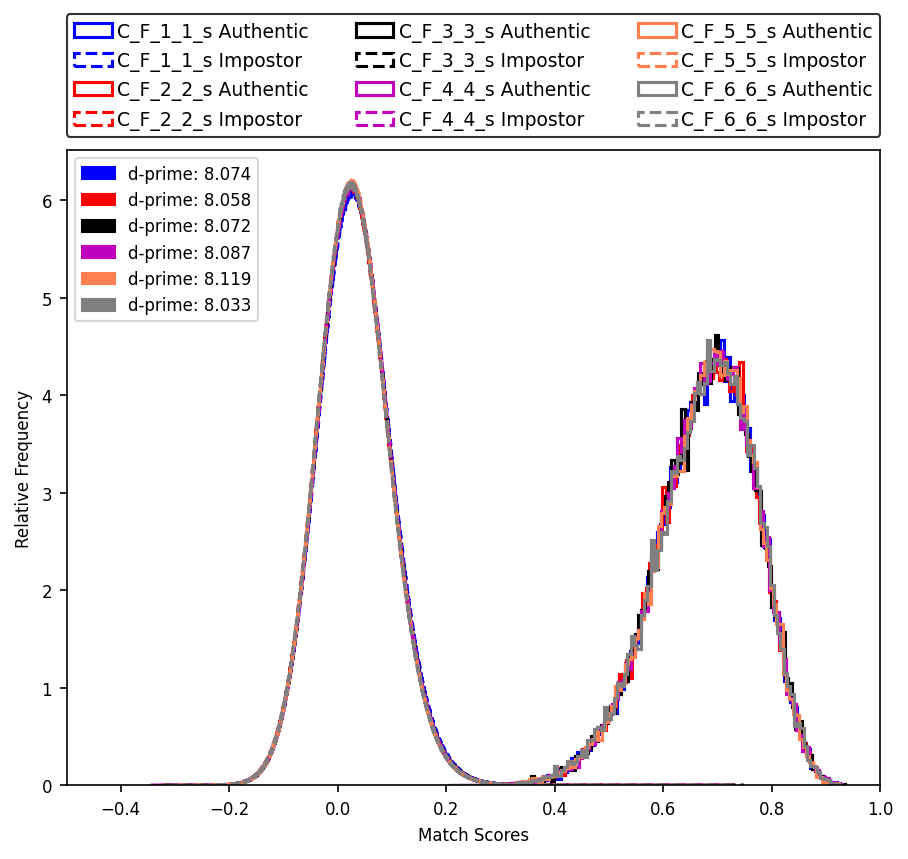}
          \caption{Resnet101}\label{r100}
      \end{subfigure}
      \hfill
      \begin{subfigure}[b]{0.33\linewidth}
        \centering
          \includegraphics[width=1\linewidth]{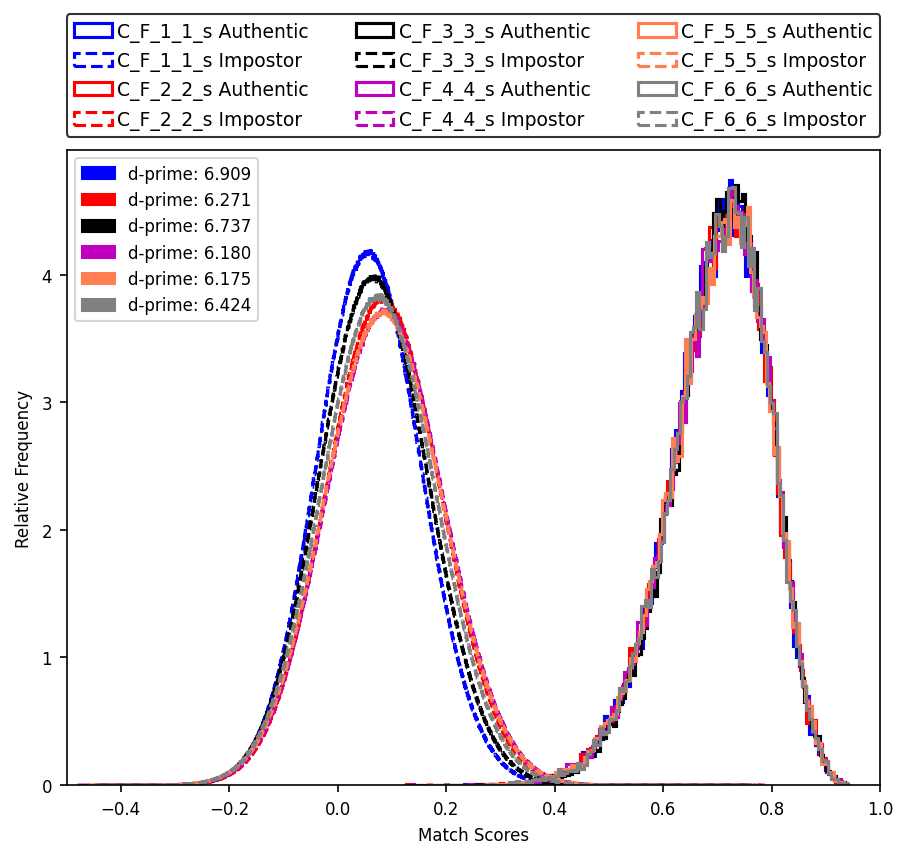}
          \caption{ViT}\label{vit}
      \end{subfigure}
  \end{subfigure}
 \caption{{\bf Training $N$ matchers with equivalent recognition power.} While $N$ instances of deep CNN models can be trained with consistent performance, ViT exhibits variations across different instances of trained models. Note that not only are the d-primes consistent across the deep CNN models, but the genuine and impostor distributions also lie closely together along the similarity axis.
}
  \vspace{-0.5em}
  \label{fig:n-models}
\end{figure*}
\begin{figure*}[t]
\centering
  \begin{subfigure}[b]{1\linewidth}
    \centering
      \begin{subfigure}[b]{0.33\linewidth}
        \centering
          \includegraphics[width=1\linewidth]{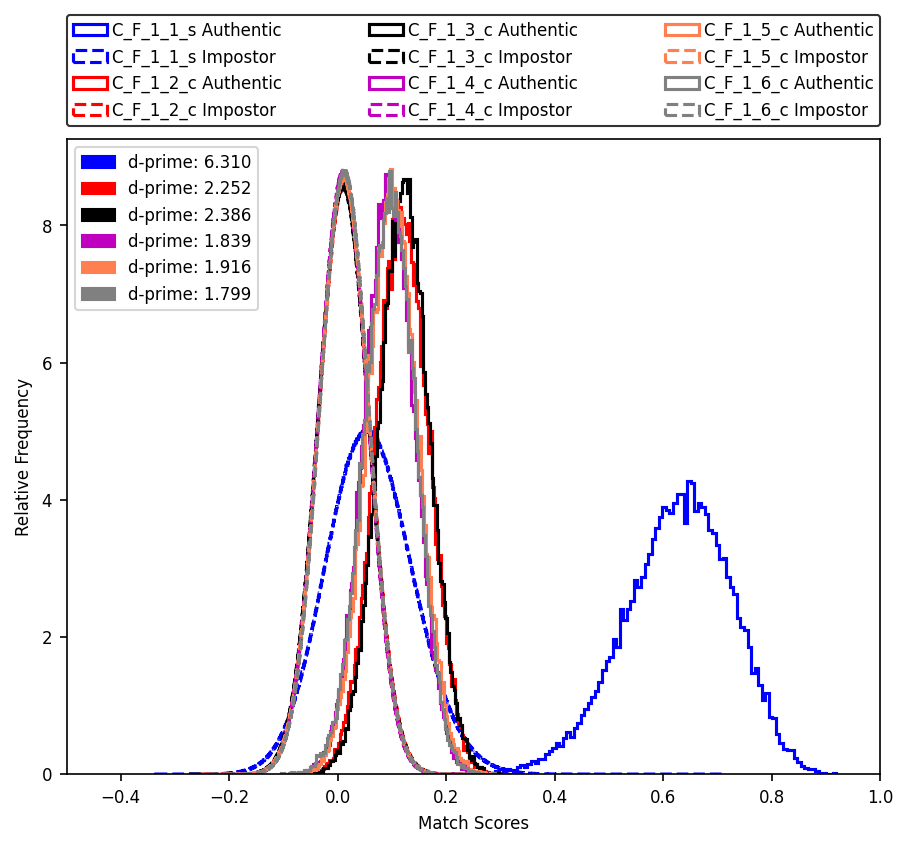}
          \caption{R18}\label{r18}
      \end{subfigure}
      \hfill
      \begin{subfigure}[b]{0.33\linewidth}
        \centering
          \includegraphics[width=1\linewidth]{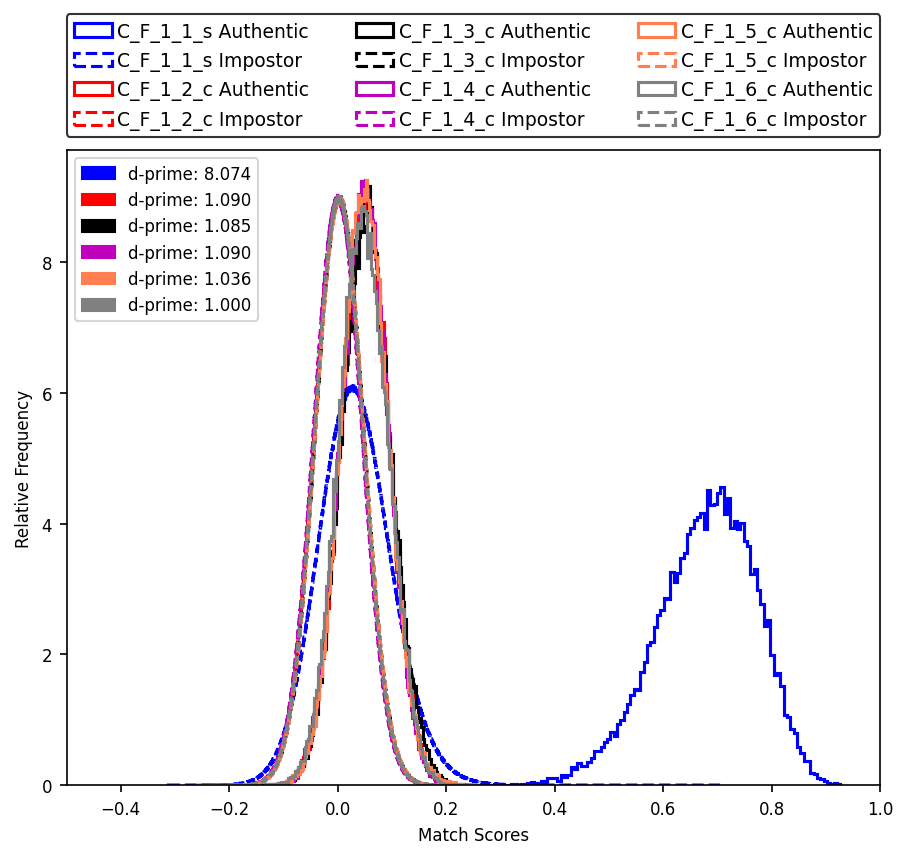}
          \caption{R100}\label{r100}
      \end{subfigure}
      \hfill
      \begin{subfigure}[b]{0.33\linewidth}
        \centering
          \includegraphics[width=1\linewidth]{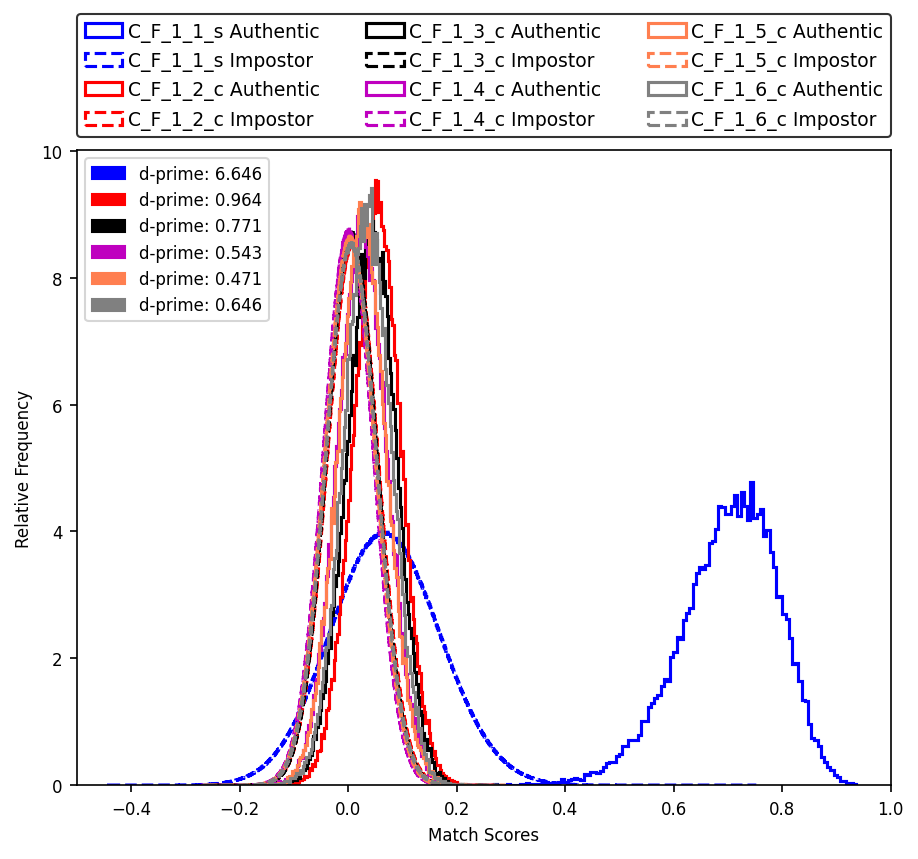}
          \caption{ViT}\label{vit}
      \end{subfigure}
  \end{subfigure}
 \caption{{\bf Cross-model genuine comparison results yield impostor-like scores.} The upper tail of the cross-model genuine distribution falls below that of the same-model genuine distribution, meaning the maximum genuine score from cross-model comparisons is lower than the maximum impostor score from same-model comparisons. This implies that the 1-in-10,000 FMR threshold used for enrollment can reliably be applied for verification, even if the model changes for identities whose stored template has been compromised and revoked. This behavior is consistent for both ResNet and ViT networks.}
  \vspace{-0.5em}
  \label{fig:cross-model}
\end{figure*}

The results in Table \ref{tab:n-models} show a low standard deviation across the benchmark datasets for ResNet networks, indicating that training multiple ResNet models with similar accuracy is feasible. However, face recognition (FR) models using Vision Transformers (ViT) as the backbone exhibit significant variation in average performance, suggesting that training \(N\) models with equal performance using ViT is comparatively more challenging. The d-prime values for each demographic group in the MORPH dataset further confirm the consistent performance across groups for ResNet backbones, while highlighting the variability in ViT backbones. Additionally, the distributions of genuine and impostor scores, shown in Figure \ref{fig:n-models}, remain consistent across the $N$ ResNet models, whereas the variations in the ViT models are more pronounced. This reinforces the reliability of face recognition networks with ResNet backbones across demographic variations for the $N$-model revocable framework proposed in this work, whereas ViT backbones do not ensure equal accuracy across all $N$ trained models.

The performance variability in Vision Transformers (ViTs) compared to ResNets can be attributed to several factors, such as the lack of image-specific inductive biases, the high sensitivity of self-attention to initialization, a more complex optimization landscape, and greater data requirements \cite{zheng_arxiv_2024}. Therefore, the ViT architecture may not be the most suitable backbone for an $N$-model revocable biometrics framework proposed in this work.

\setcounter{figure}{0}
\begin{figure*}[t]
    \centering
    \includegraphics[width=0.9\textwidth]{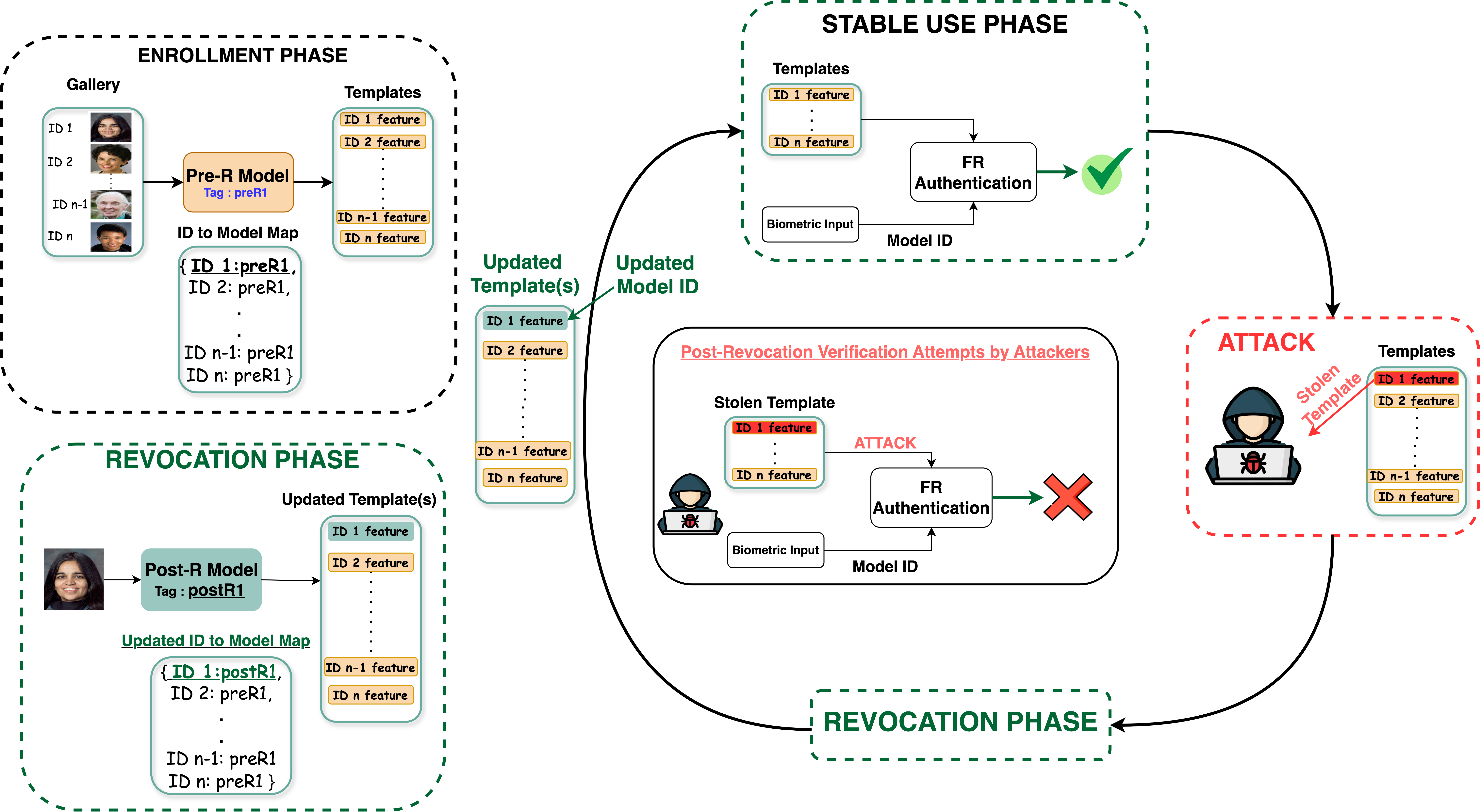}
    \caption{{\bf Proposed Revocable Biometric System Flow: Enrollment, Revocation, and Update. } The proposed biometric system begins with the enrollment phase. Once enrolled, the system operates stably. If a template is stolen or a user requests revocation, the revocation phase is initiated, during which the existing template is updated using the next available instance of the trained model, and the identity-to-model mapping is also updated. This process renders the old template unusable, allowing the system to resume stable operation.}
    \label{fig:overall_framework}
    \vspace{-0.75em}
\end{figure*}
\subsection{Template Utility Only ``Within the Model"}
One of the key principles in designing a revocable biometric system using \(N\) distinct trained models is that feature vectors from genuine pairs, when extracted using different model instances, should yield low cosine similarity scores, causing genuine pairs to behave like impostor pairs. This characteristic ensures that revoked templates cannot be used to impersonate the legitimate user after re-enrollment.

To illustrate that cross-model genuine pairs produce impostor-like scores, we use six instances of the same end-to-end model from Section \ref{nmodels}. One model is randomly selected as the reference, with the remaining five serving as alternative models. For the cross-model comparison, we extract the embedding of one image of a genuine pair using the reference model and the embedding for the other image from the pair using one of the alternative models. The genuine and impostor score distributions from these cross-model feature match comparisons are shown in Figure \ref{fig:cross-model}. Two key observations can be made from the plots. First, the cross-model comparison typically results in a much lower d-prime than the same-model comparison. Second, not only is the d-prime lower, but both the genuine and impostor distributions in the cross-model comparisons are shifted toward a significantly lower score range. {\it More importantly, the upper tail of the cross-model genuine distribution falls below that of the same-model impostor distribution, meaning that even the highest match score from the cross-model genuine comparison is lower than the highest impostor score from the same-model comparison.}

This holds largely true for both ResNets and ViT models. However, as discussed in Section \ref{nmodels}, ViT exhibits significant variability in model performance across $N$ training instances. If methods to stabilize ViT training are developed, face recognition models with ViT backbones could also be used within this framework, as cross-model genuine matches would behave like impostors, similar to ResNets models.

\subsection{System Overview}

Given that we can train \(N\) equally capable models using the same architecture, training datasets, and hyperparameter configurations, and that cross-model comparisons yield impostor-like scores for genuine pairs, we can develop a revocable biometric system that leverages multiple models to effectively counter impostor attacks. To illustrate this concept, let's assume we have \(N\) sets of trained models available for use. For clarity, we refer to the model that is currently active as the {\bf Pre-Revocation Model (pre-R)}. In the event of a security breach or if the enrollee decides to revoke their previous template, it is assumed that the attacker would have accessed the template generated by this model. When a template for a particular identity is revoked and a new model is assigned, this new model is referred to as the {\bf Post-Revocation Model (post-R)} or the ``secure model." This approach ensures that, even if a template is stolen or compromised, the compromised enrollment can be effectively revoked without affecting other existing enrollments. The individual's enrollment is then updated with the new model for future matching. If multiple breaches occur for the same identity, the process of updating the currently active model instance (pre-Revocation model) to a newer instance (post-Revocation model) can be repeated as needed.

The proposed system overview is shown in Figure \ref{fig:overall_framework} and operates as follows:
\begin{enumerate}
    \item {\bf Enrollment Phase:} An identity is enrolled using, say, the Pre-Revocation Model (pre-R), which generates and stores feature vectors for that identity. In addition to feature vector generation and template storage, an identity-to-model mapping dictionary is created. This mapping enables the system administrator to determine which model to use for a particular identity during a verification request. The identity-to-model mapping and update is handled through a simple hashmap lookup process, with overhead that is almost negligible. 

    \item {\bf Stable Use Phase:} While there has been no breach and no enrollee has requested the revocation of their previous template, the system remains in a stable use phase. During this phase, when a verification request is made using identity-specific biometric input, the system retrieves the designated model for that identity, generates the feature using the model, and compares the generated feature with the enrolled template.
    
    \item {\bf Revocation Phase:} If a template for a particular identity is stolen or compromised, the compromised enrollment can be revoked by invalidating the Pre-Revocation Model (pre-R) for that identity. The update process is straightforward. A different instance of the trained model, now designated as the Post-Revocation Model (post-R), is selected from one of the $N$ available trained models and assigned to the compromised identity. The gallery template for the identity is updated using features extracted by the Post-Revocation Model. The identity-to-model mapping in the dictionary is then updated to reflect this change. This process can be applied to multiple identities in cases where several identities request revocation or multiple breaches occur. By doing this, the stolen or compromised template can no longer be used for authentication. The new model generates fresh feature vectors that are unrelated to the previous compromised model. {\it Note that re-enrolling compromised identities using the new Post-Revocation (post-R) model does not require re-enrollment for all other identities in the gallery}. One minor overhead of this system is the need to maintain a mapping of each identity to the specific model they are enrolled with, which is required for future verification requests. This process of revocation can be repeated an unlimited number of times, ensuring continued usage in the case of multiple breaches.
    
    \item {\bf Return to Stable Use Phase:} Given that cross-model comparisons result in impostor-like matches, when a verification request involves an identity whose template has been breached or revoked, future authentications will be handled by the Post-Revocation Model (post-R) and an updated template gallery generated using this model for the specific or multiple affected identities. This ensures that any templates accessed by malicious actors are treated as impostor matches and denied access, while legitimate users can continue to use the system with the same accuracy as the previous model, despite the revocation of their earlier template. With updated identity-to-model mapping and an updated template gallery, the system can return to the stable use phase until the next breach event. 
\end{enumerate}

\section{Implementation Details}

To implement our re-enrollment schemes for comparison experiments, we adopt ResNet18 and ResNet100 \cite{he_cvpr_2016deep} with the modifications proposed in \cite{deng_cvpr_2009} and ViT as detailed in \cite{Insightface}. We use ArcFace loss as the choice of our loss function, with combined margin values of (1, 0, 0.4). We use WebFace4M dataset \cite{zhu_cvpr_2021} as the training set. Images in WebFace4M dataset are pre-aligned using RetinaFace \cite{deng_cvpr_2020retinaface}. The model is trained for 20 epochs using SGD as the optimizer \cite{paszke_nips_2019}, with momentum of 0.9, an initial learning rate of 0.1 and weight decay of 5e-4. We adopt polynomial decay as the learning rate scheduler during training from \cite{Insightface}. All the mentioned configuration parameters align with the ones utilized for training WebFace4M on the ResNet-50/100 backbone, as mentioned in insightface \cite{Insightface} GitHub repository. Each instance of the model (smallest to largest) requires about 10-18 hours of training on 4 RTX-6000 GPUs. 

% \begin{figure*}[t]
% \centering
%   \begin{subfigure}[b]{1\linewidth}
%     \centering
%       \begin{subfigure}[b]{0.33\linewidth}
%         \centering
%           \includegraphics[width=1\linewidth]{images/final-results/C_F_r18_arcface.png}
%           \caption{R18}\label{r18-final}
%       \end{subfigure}
%       \hfill
%       \begin{subfigure}[b]{0.33\linewidth}
%         \centering
%           \includegraphics[width=1\linewidth]{images/final-results/C_F_r100_arcface.png}
%           \caption{R100}\label{r100-final}
%       \end{subfigure}
%       \hfill
%       \begin{subfigure}[b]{0.33\linewidth}
%         \centering
%           \includegraphics[width=1\linewidth]{images/final-results/C_F_vit_arcface.png}
%           \caption{ViT}\label{vit-final}
%       \end{subfigure}
%   \end{subfigure}
%  \caption{ Final Model Comparison.}
%   \vspace{-0.5em}
%   \label{fig:whole pipeline}
% \end{figure*}

\begin{figure*}[t]
\centering
  \begin{subfigure}[b]{1\linewidth}
    \centering
      \begin{subfigure}[b]{0.33\linewidth}
        \centering
          \includegraphics[width=1\linewidth]{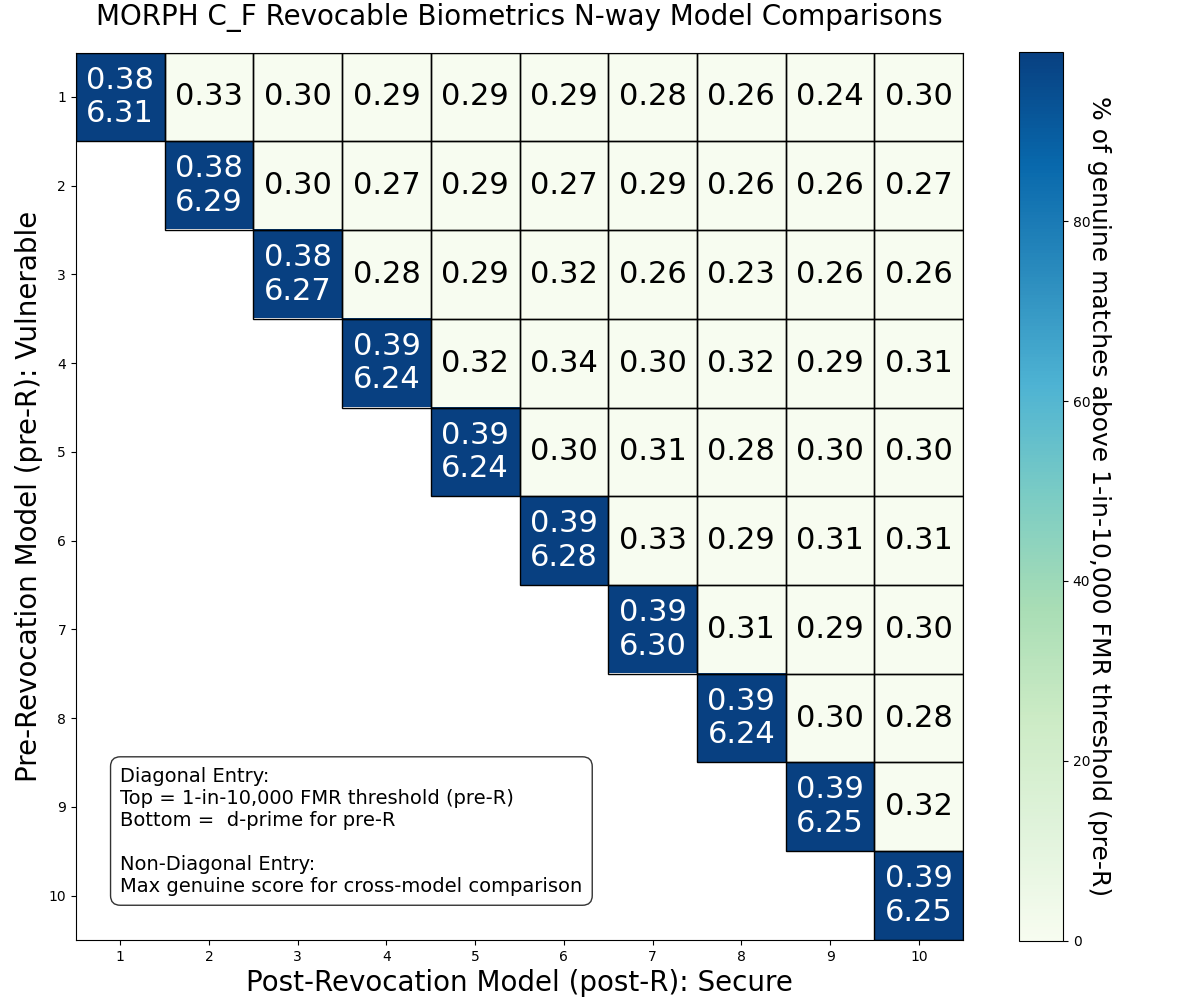}
          \caption{C F w/ ResNet18}\label{r18-final1}
      \end{subfigure}
      \hfill
      \begin{subfigure}[b]{0.33\linewidth}
        \centering
          \includegraphics[width=1\linewidth]{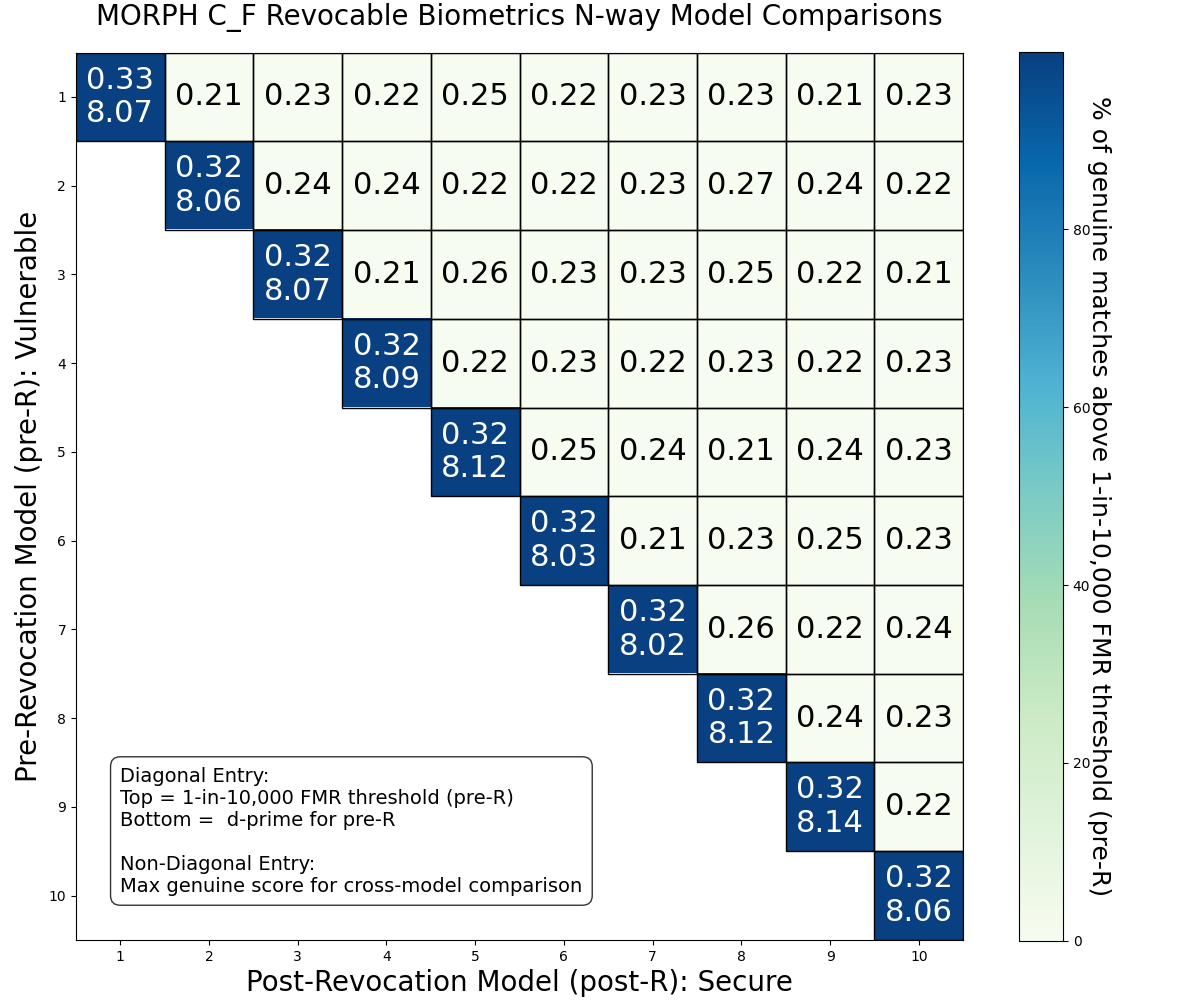}
          \caption{C F w/ ResNet100}\label{r100-final1}
      \end{subfigure}
      \hfill
      \begin{subfigure}[b]{0.33\linewidth}
        \centering
          \includegraphics[width=1\linewidth]{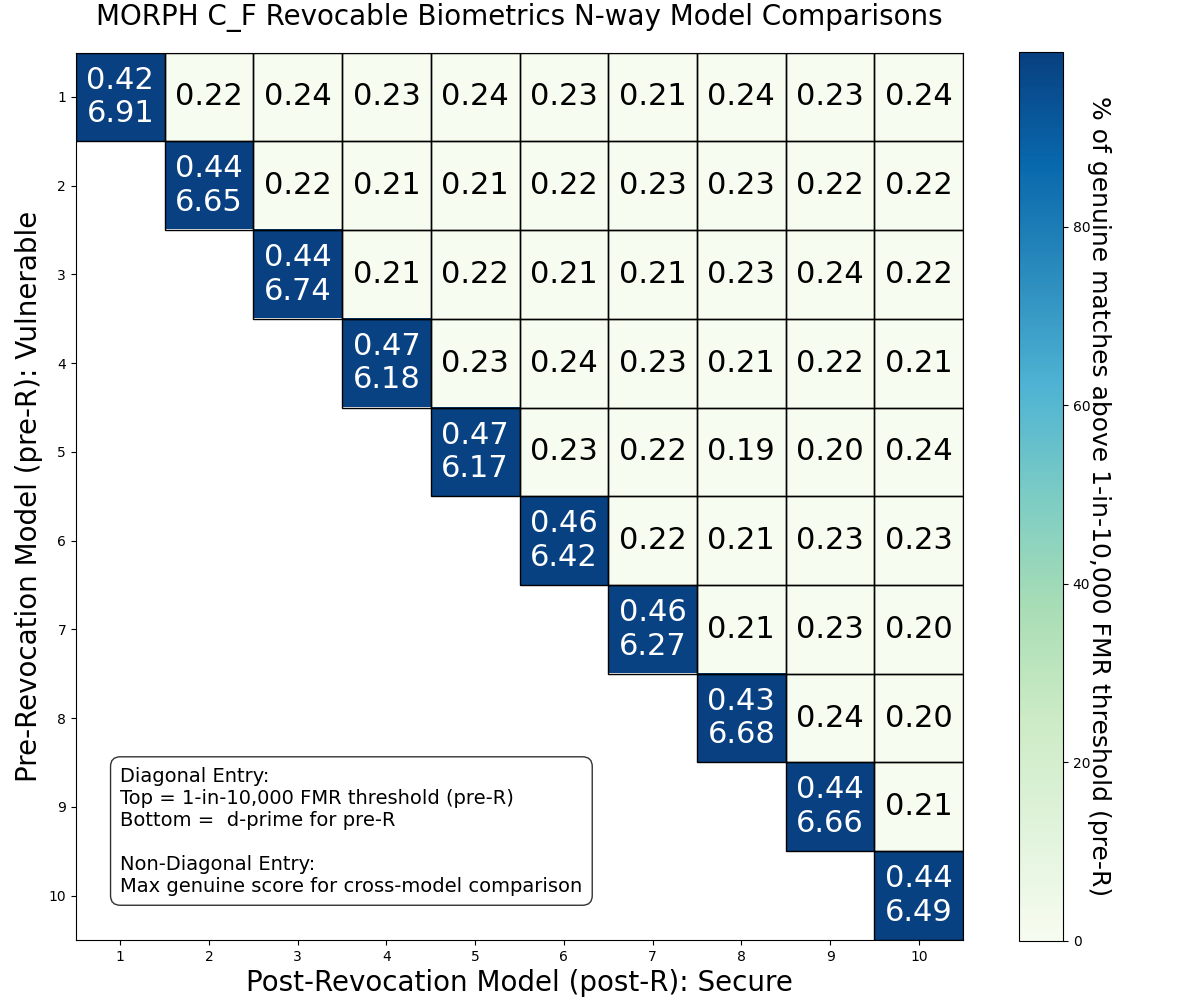}
          \caption{C F w/ ViT}\label{vit-final1}
      \end{subfigure}
  \end{subfigure}
  \hfill \vspace{0.25em}
  \begin{subfigure}[b]{1\linewidth}
    \centering
      \begin{subfigure}[b]{0.33\linewidth}
        \centering
          \includegraphics[width=1\linewidth]{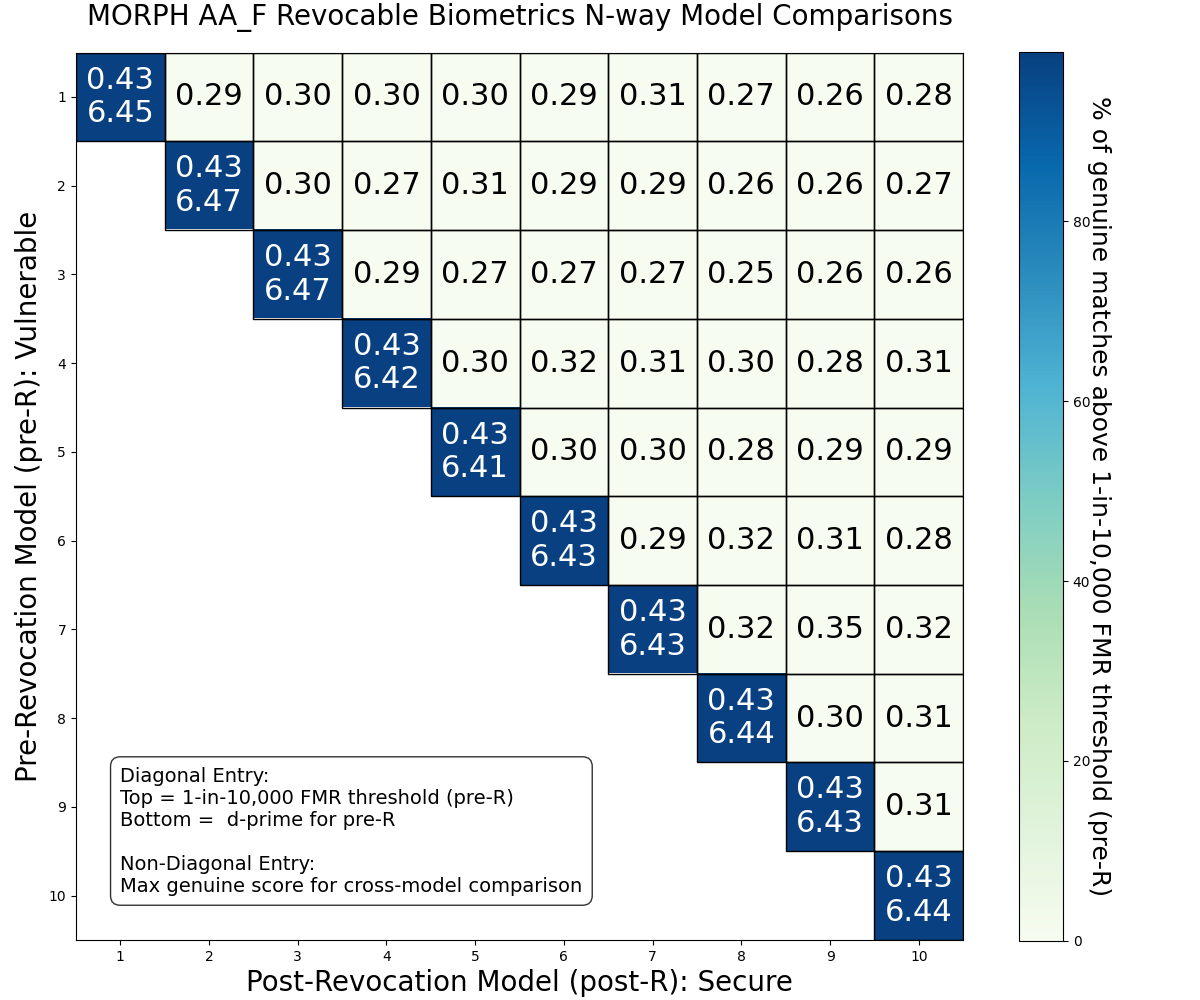}
          \caption{AA F w/ ResNet18}\label{r18-final2}
      \end{subfigure}
      \hfill
      \begin{subfigure}[b]{0.33\linewidth}
        \centering
          \includegraphics[width=1\linewidth]{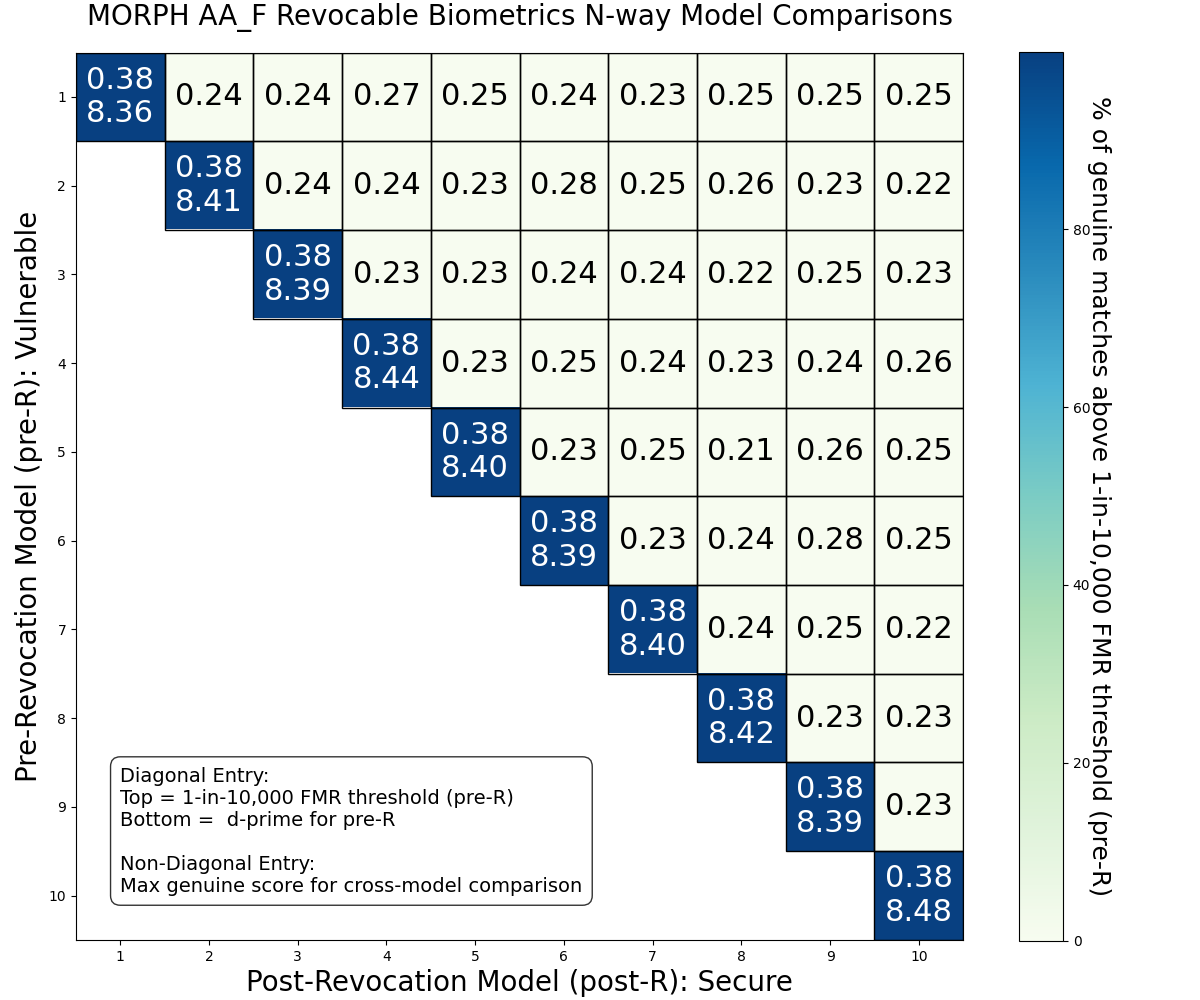}
          \caption{AA F w/ ResNet100}\label{r100-final2}
      \end{subfigure}
      \hfill
      \begin{subfigure}[b]{0.33\linewidth}
        \centering
          \includegraphics[width=1\linewidth]{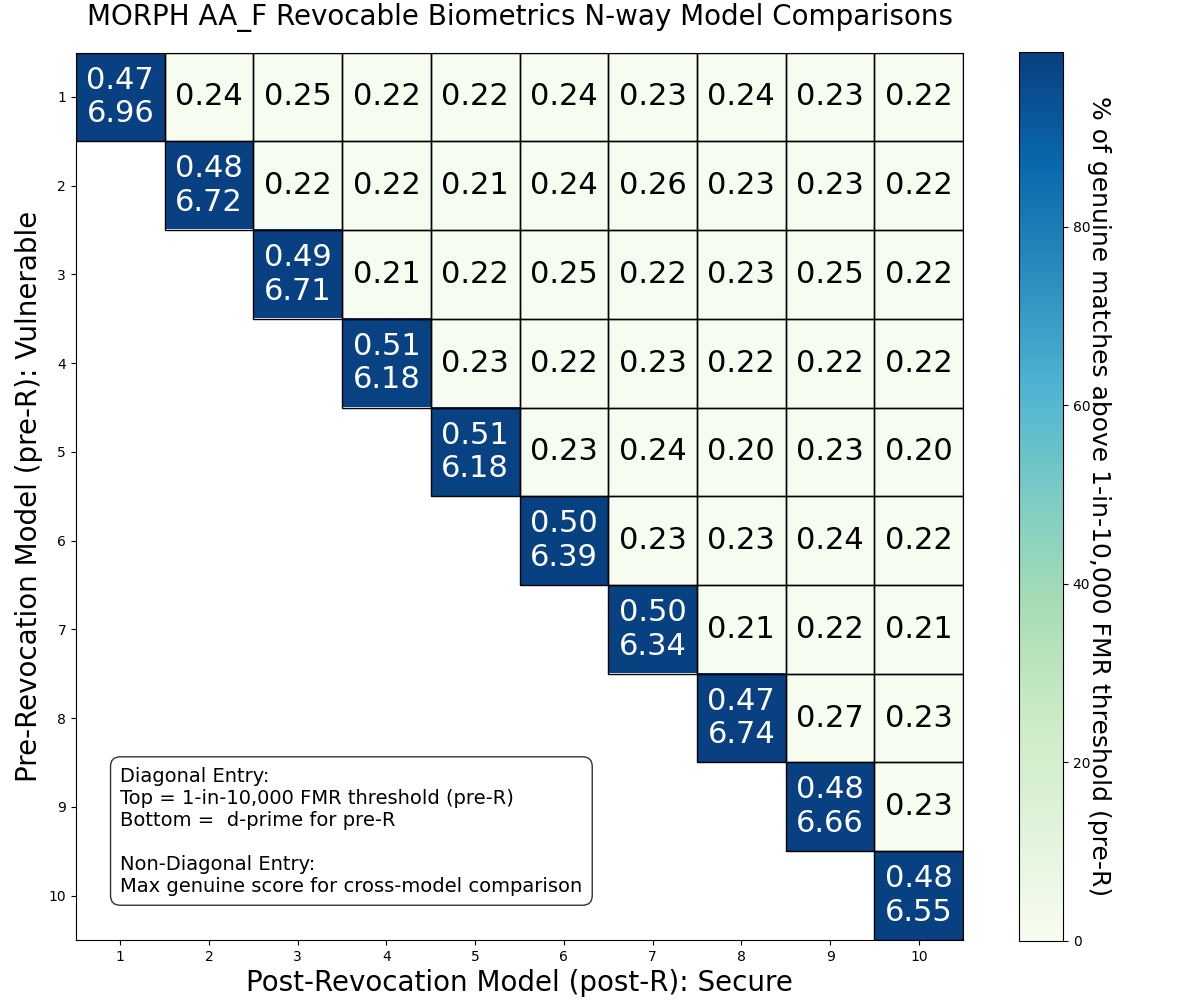}
          \caption{AA F w/ ViT}\label{vit-final2}
      \end{subfigure}
  \end{subfigure}
 \caption{ {\bf Relationship matrix illustrating that verification attempts using a revoked template are unsuccessful.} {\it Top item in diagonal entries represents the same-model 1-in-10,000 FMR threshold, the bottom item in diagonal entries represents the same-model d-prime, and non-diagonal entries represent the maximum genuine score from cross-model comparisons}. The cosine similarity scores for all 45 cross-model comparisons are below the operational threshold of 1-in-10,000 FMR for the original reference model, indicating that even the highest cross-model genuine comparisons do not meet the similarity criteria for successful verification. {\it The top row presents the results for the Caucasian Female group, while the bottom row shows the results for the African-American Female group.}}
  \vspace{-0.5em}
  \label{fig:whole pipeline}
\end{figure*}

\section{Results and Analysis}

To demonstrate the results of our proposed system, we selected the Caucasian Female and African American Female groups from the MORPH dataset. These groups are generally considered to exhibit lower accuracy compared to other demographic groups, making them ideal for showcasing the robustness of our proposed revocable biometric system. To demonstrate the effectiveness of our results, we trained 10 instances of the same end-to-end model. With 10 models, there are \( n \times (n-1) / 2 \) unique model pairs, resulting in a total of 45 pairs. We present both same-model and cross-model comparisons using a relationship matrix structure. {\it The cross-model comparisons here represent an attacker supplying a stolen template, which is then compared to a re-enrolled template after the user revokes and re-enrolls.} The diagonal entries represent same-model comparisons (both images in the pair use the same model), while the off-diagonal entries show cross-model comparisons (two different models are used for the images in a pair). Since feature matching is commutative, the relationship matrix is symmetric, with cross-model comparisons shown in the upper triangle of the matrix. \\

\noindent{\bf Same model probe to gallery comparison.} The diagonal entries in Figure \ref{fig:whole pipeline} represent standard feature matching between the probe and the feature template generated from the same model. {\it In each diagonal cell, the top value indicates the d-prime of the model, while the bottom value indicates the 1-in-10000 FMR threshold}. The d-prime serves as a general performance indicator for the model, and the 1-in-10,000 FMR threshold represents the lower bound match value for genuine authentication. From Figures \ref{r18-final1}-\ref{r100-final1}, and  \ref{r18-final2}-\ref{r100-final2}, we observe that across all diagonal entries, the d-prime and 1-in-10,000 FMR thresholds for ResNet networks are remarkably consistent. This demonstrates that it is empirically possible to train a large number of models with similar performance, allowing for continuous template revocation and re-enrollment with a new template. {\it From Figure \ref{fig:n-models}, the position of the impostor and genuine overlaps across $N$ models indicates that the same threshold can be applied to the newer post-Revocation model.} In contrast, Figures \ref{vit-final1} and \ref{vit-final2} show considerable variation in d-prime across 10 different instances of the ViT-trained models. This suggests that ResNet-based networks are inherently better suited for generating multiple models with similar performance compared to ViT-based networks. \\

\noindent{\bf Cross model probe to gallery comparison.} The non-diagonal entries in Figure \ref{fig:whole pipeline} represent standard feature matching between the probe and the feature template generated from different models. {\it Each non-diagonal entry shows the maximum genuine score produced by cross-model comparisons}. For instance, the entry at cell (1,2) represents the highest genuine match score obtained when using the probe feature extracted by model 1 and the gallery template extracted by model 2. In an operational context, if a template for any identity is compromised, the gallery feature for that identity is re-extracted using the post-revocation model (represented by all the column entries).

From Figures \ref{r18-final1} - \ref{vit-final2}, we observe that the non-diagonal entries, representing the cross-model maximum genuine scores, are all lower than the operational 1-in-10,000 FMR. For example, if model 1 (the top-left entry) is the pre-Revocation model, the 1-in-10,000 FMR for this model is 0.38. This value represents the lower bound of the match score, meaning that for any biometric sample to be declared a genuine match, its similarity to the stored template must exceed this threshold. However, for model 1 (top row across all columns), the maximum genuine score for all cross-model comparisons between probe and gallery features is lower than this threshold. This indicates that even the best possible genuine scores using different models for probe and gallery are below this limit, ensuring that even if an attacker has access to the gallery template from the revoked model, the system will still reject the match as genuine.

Although training $N$ equally accurate instances of ViT-based models is challenging, and ViTs are not yet state-of-the-art in face recognition, their cross-model matching behavior is very similar to that of CNNs. This suggests that, if ViT-based networks can be trained without performance variance, they could reliably be used in the $N$-model revocable biometric system proposed in this work.

\section{Conclusion and Discussion}

We describe a general approach to revocable biometrics for deep CNN based face recognition.  Our approach exploits the fact that multiple trainings of a ResNet-based face matcher result in matchers that have equivalent accuracy, yet those matchers also generate embeddings for a given face image that are incompatible across matchers.  The result is that if a given user’s enrolled template is compromised in some way, they can it revoked and be re-enrolled, potentially an unlimited number of times, without experiencing any degradation in accuracy.

Our approach requires that a user’s face image be available for re-enrollment.  This can either be an archived image, if policies allow this, or the user can present a fresh image as part of revocation and re-enrollment.  This is not an onerous requirement, as in current industry practice, a new release of a matcher can require computing new templates for all enrolled images.  Our approach also requires that the system maintain a list of which model instance is currently valid for each user.  The system can conceivably control or limit the number of model instances, if desired, by revoking old templates and re-enrolling with a newer model instance.   Our approach also requires training multiple instances of a matcher.  These can be done in batches ahead of them being needed, so that a request for revocation and re-enrollment can be satisfied immediately, or a new instance can be trained as needed when a request for revocation and re-enrollment is made, resulting in a short time before the re-enrollment is completed.

Revocable face recognition does not alleviate the need for strong presentation attack detection (PAD).  Revocability and strong PAD are complementary elements of a secure face recognition system.  PAD has received greater attention because in modern society basically everyone has multiple of their face images in the public domain and accessible to hackers.  Presentation Attack Detection (PAD) is the solution to malicious actors attempting to impersonate a targeted individual using their face image.  Revocability is the solution to the deeper problem that a malicious actors has stolen either the enrolled template associated with a person or a fresh template created when the person makes an identity verification transaction, and is using the stolen template to impersonate the person.

\noindent{\bf Future Work.} There are various questions that could be addressed in future work.  One is that our experimental results in this paper are based on ResNet deep CNNs, and it would be useful to verify that other popular CNNs for face recognition support revocability equally well.  An interesting theoretical question involves the maximum number of equally accurate but distinct models that a network can generate.  Empirically it is clear that the number is large for practical purposes, but it would be interesting to know the theoretical upper limit.  Another interesting question is – what is the minimum training effort needed to generate an equally accurate but distinct model.  In this work, we have done complete trainings from scratch.  But it is possible that a fine-tuning step that involves less computation could still produce a suitable model.   Future research could also investigate the unreliability of ViT backbones in cancellable biometric applications and try to identify conditions under which ViTs can be effectively employed.  Lastly, future research could explore whether this approach to revocable biometrics applies to other modalities, such as iris and fingerprint, where CNNs are used as the feature extractors.

\section*{Ethical Impact Statement}
\noindent This work aims to improve the security of face recognition systems against attacks by utilizing the inherent capabilities of deep CNN matchers to design a secure, revocable biometric system. Our results demonstrate that this revocable framework performs consistently across all demographics considered. No human data was directly collected for the experimental results presented in this paper.

{\small
\bibliographystyle{ieee}
\bibliography{egbib}
}

\end{document}